\pdfoutput=1

\documentclass[10pt,twocolumn,letterpaper]{article}

\usepackage{cvpr}
\usepackage{times}
\usepackage{epsfig}
\usepackage{graphicx}
\usepackage{amsmath}
\usepackage{amssymb}

\usepackage{enumerate}
\usepackage{algorithm}
\usepackage{algorithmic}

\usepackage[breaklinks=true,bookmarks=false]{hyperref}

\cvprfinalcopy 

\def\cvprPaperID{7176} 

\ifcvprfinal\fi
\begin{document}

\title{SLV: Spatial Likelihood Voting for Weakly Supervised Object Detection}


\author{Ze Chen\textsuperscript{1,2,}\footnotemark[1] \hspace{0.3cm} Zhihang Fu\textsuperscript{5} \hspace{0.3cm} Rongxin Jiang\textsuperscript{1,3} \hspace{0.3cm} Yaowu Chen\textsuperscript{1,4,}\footnotemark[2] \hspace{0.3cm} Xian-sheng Hua\textsuperscript{5,}\footnotemark[2] \\
\textsuperscript{1}{Zhejiang University, Institute of Advanced Digital Technology and Instrument}\\
\hspace{-0.05cm}\textsuperscript{2}{Zhejiang University Embedded System Engineering Research Center, Ministry of Education of China}\\
\textsuperscript{3}{Zhejiang University, the State Key Laboratory of Industrial Control Technology}\\
\textsuperscript{4}{Zhejiang Provincial Key Laboratory for Network Multimedia Technologies}\\
\textsuperscript{5}{Alibaba DAMO Academy, Alibaba Group}\\
{\tt\small \{chenze,rongxinj\}@zju.edu.cn \{zhihang.fzh,xiansheng.hxs\}@alibaba-inc.com}\\
{\tt\small cyw@mail.bme.zju.edu.cn}
}

\maketitle
\thispagestyle{empty}

\renewcommand{\thefootnote}{\fnsymbol{footnote}}
\footnotetext[1]{This work was done when the author was visiting Alibaba as a research intern.}
\footnotetext[2]{Corresponding authors.}
\begin{abstract}
   Based on the framework of multiple instance learning (MIL), tremendous works have promoted the
   advances of weakly supervised object detection (WSOD). 
   However, most MIL-based methods tend to localize instances to their discriminative parts instead of the whole content.
   In this paper, we propose a spatial likelihood voting (SLV) module to converge the proposal
   localizing process without any bounding box annotations.
   Specifically, all region proposals in a given image play the role of voters every iteration during training, voting for the likelihood of each category in spatial dimensions. 
   After dilating alignment on the area with large likelihood values, the voting results are regularized as bounding boxes, being used for the final classification and localization. 
   Based on SLV, we further propose an end-to-end training framework for multi-task learning.
   The classification and localization tasks promote each other, which further improves the detection performance.   
   Extensive experiments on the PASCAL VOC 2007 and 2012 datasets demonstrate the superior performance of SLV.
\end{abstract}

\section{Introduction}

Object detection is an important problem in computer vision, which aims at localizing tight bounding boxes
of all instances in a given image and classifying them respectively. 
With the development of convolutional neural network (CNN)~\cite{he2016deep,krizhevsky2012imagenet,lecun1998gradient}
and large-scale annotated datasets~\cite{everingham2015pascal,lin2014microsoft,russakovsky2015imagenet}, there have
been great improvements in object detection~\cite{girshick2015fast,girshick2014rich,lin2017feature,liu2016ssd,ren2015faster}
in recent years. However, it is time-consuming and labor-intensive to annotate accurate object bounding boxes
for a large-scale dataset. Therefore, weakly supervised object detection (WSOD), which only use image-level
annotations for training, is considered to be a promising solution in reality and has attracted the attention
of academic community in recent years.

\begin{figure}[t]
\begin{center}
   \includegraphics[width=1.0\linewidth]{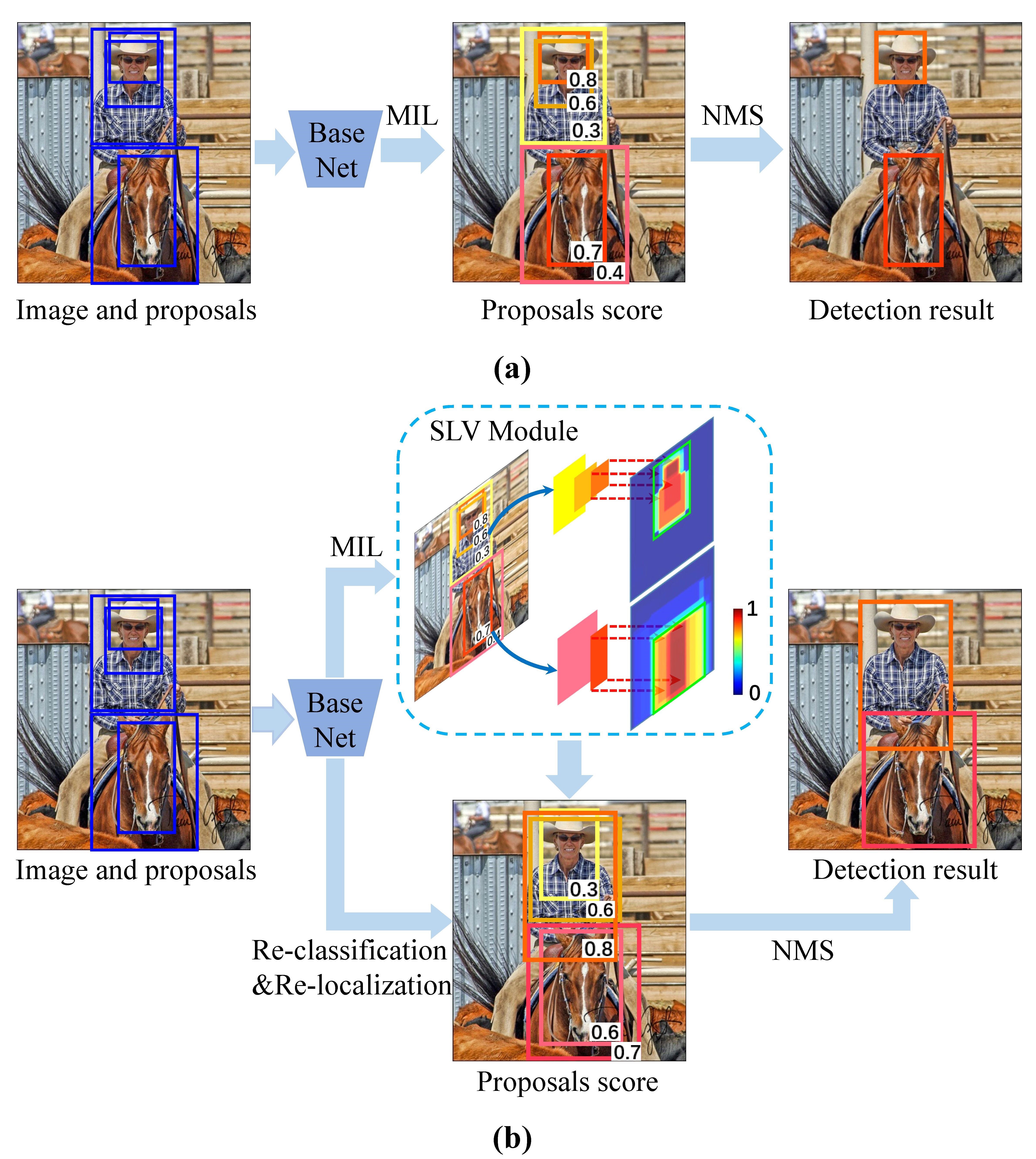}
\end{center}
   \caption{Detection results without/with SLV module.
   (a) Common MIL-based methods are easy to localize instances to their discriminative parts instead of the whole content.
   (b) SLV module shifts object proposals and detects accurate bounding boxes of objects.}
\label{fig:long}
\label{fig:fig1}
\end{figure}

Most WSOD methods~\cite{bilen2016weakly,cinbis2016weakly,ren2015weakly,tang2018pcl,tang2017multiple,wei2018ts2c}
follow the multiple instance learning (MIL) paradigm.
Regarding the WSOD as an instance classification problem, they train an instance classifier under MIL constraints to approach the purpose of object detection.
However, the existing MIL-based methods only focus on feature representations for instance classification without
considering localization accuracy of the proposal regions. As a consequence, they tend to localize instances to
their discriminative parts instead of the whole content, as illustrated in Fig~\ref{fig:fig1}(a).

Due to lack of bounding box annotations, the absence of localization task is always a serious problem in WSOD.
As a remedy, the subsequent works~\cite{li2016weakly,tang2018pcl,tang2017multiple,wan2019c} choose to re-train a
Fast-RCNN~\cite{girshick2015fast} detector fully-supervised with pseudo ground-truths, which are generated by
MIL-based weakly-supervised object detectors. The fully-supervised Fast-RCNN alleviates the above-mentioned
problem by means of multi-task training~\cite{girshick2015fast}. But it is still far from the optimal solution.

In this paper, we propose a spatial likelihood voting (SLV) module to converge the proposal localizing process without any bounding box annotations. The spatial likelihood voting operation consists of instances selection, spatial probability accumulation, and high likelihood region voting. 
Unlike the previous methods which always
keep the position of their region proposals unchanged, all region proposals in SLV play the role of voters every
iteration during training, voting for the likelihood of each category in spatial dimensions. Then the voting
results, which will be used for the re-classification and re-localization shown in Fig~\ref{fig:fig1}(b), are
regularized as bounding boxes by dilating alignment on the area with large likelihood values. Through generating
the voted results, the proposed SLV evolves the instance classification problem into multi-tasking field. SLV opens
the door for WSOD methods to learn classification and localization simultaneously. Furthermore, we propose an
end-to-end training framework based on SLV module. The classification and localization tasks promote each other,
which finally educe better localization and classification results and shorten the gap between weakly-supervised
and fully-supervised object detection. 

In addition, we conduct extensive experiments on challenging PASCAL VOC datasets~\cite{everingham2015pascal} to
confirm the effectiveness of our method. The proposed framework achieves 53.5\% and 49.2\% mAP on VOC 2007 and
VOC 2012 respectively, which, to the best of our knowledge, is the best single model performance to date.

The contributions of this paper are summarized as follows:
\begin{enumerate}[1)]
\item We propose a spatial likelihood voting (SLV) module to converge the proposal localizing process with only
image-level annotations.The proposed SLV evolves the instance classification problem into multi-tasking field.
\item We introduce an end-to-end training strategy for the proposed framework,
which boosts the detection performance by feature representation sharing.
\item Extensive experiments are conducted on different datasets.
The superior performance suggests that a sophisticated localization fine-tuning should be a promising
exploration in addition to the independent Fast-RCNN re-training.
\end{enumerate}

\section{Related Work}
MIL is a classical weakly supervised learning problem and now is a major approach
to tackle WSOD. MIL treats each training image as a “bag” and candidate proposals
as “instances”. The objective of MIL is to train an instance classifier to select
positive instances from this “bag”. With the development of the Convolution Neural Network,
many works~\cite{bilen2016weakly,diba2017weakly,kantorov2016contextlocnet,tang2017deep}
combine CNN and MIL to deal with the WSOD problem. For example, Bilen and
Vedaldi~\cite{bilen2016weakly} propose a representative two-stream weakly supervised
deep detection network (WSDDN), which can be trained with image-level annotations in
an end-to-end manner. 
Based on the architecture in~\cite{bilen2016weakly},~\cite{kantorov2016contextlocnet}
proposes to exploit the contextual information from regions around the object as a
supervisory guidance for WSOD.

In practice, MIL solutions are found easy to converge to discriminative parts of
objects. This is caused by the loss function of MIL is non-convex and thus MIL
solutions usually stuck into local minima. To address this problem, Tang\etal~\cite{tang2017multiple} combine
WSDDN with multi-stage classifier refinement and propose an OICR algorithm to help their network see larger
parts of objects during training. Moreover, building on~\cite{tang2017multiple}, Tang\etal~\cite{tang2018pcl}
subsequently introduce the proposal cluster learning and use the proposal clusters as supervision which
indicates the rough locations where objects most likely appear. In~\cite{wan2018min}, Wan \etal try to reduce
the randomness of localization during learning. In~\cite{zhang2018zigzag}, Zhang \etal add curriculum learning
using the MIL framework. From the perspective of optimization, Wan \etal~\cite{wan2019c} introduce the continuation
method and attempt to smooth the loss function of MIL with the purpose of alleviating the
non-convexity problem. In~\cite{gao2019utilizing}, Gao \etal make use of the instability of
MIL-based detectors and design a multi-branch network with orthogonal initialization.

\begin{figure*}
\begin{center}
\includegraphics[width=5.9in,height=2.83in]{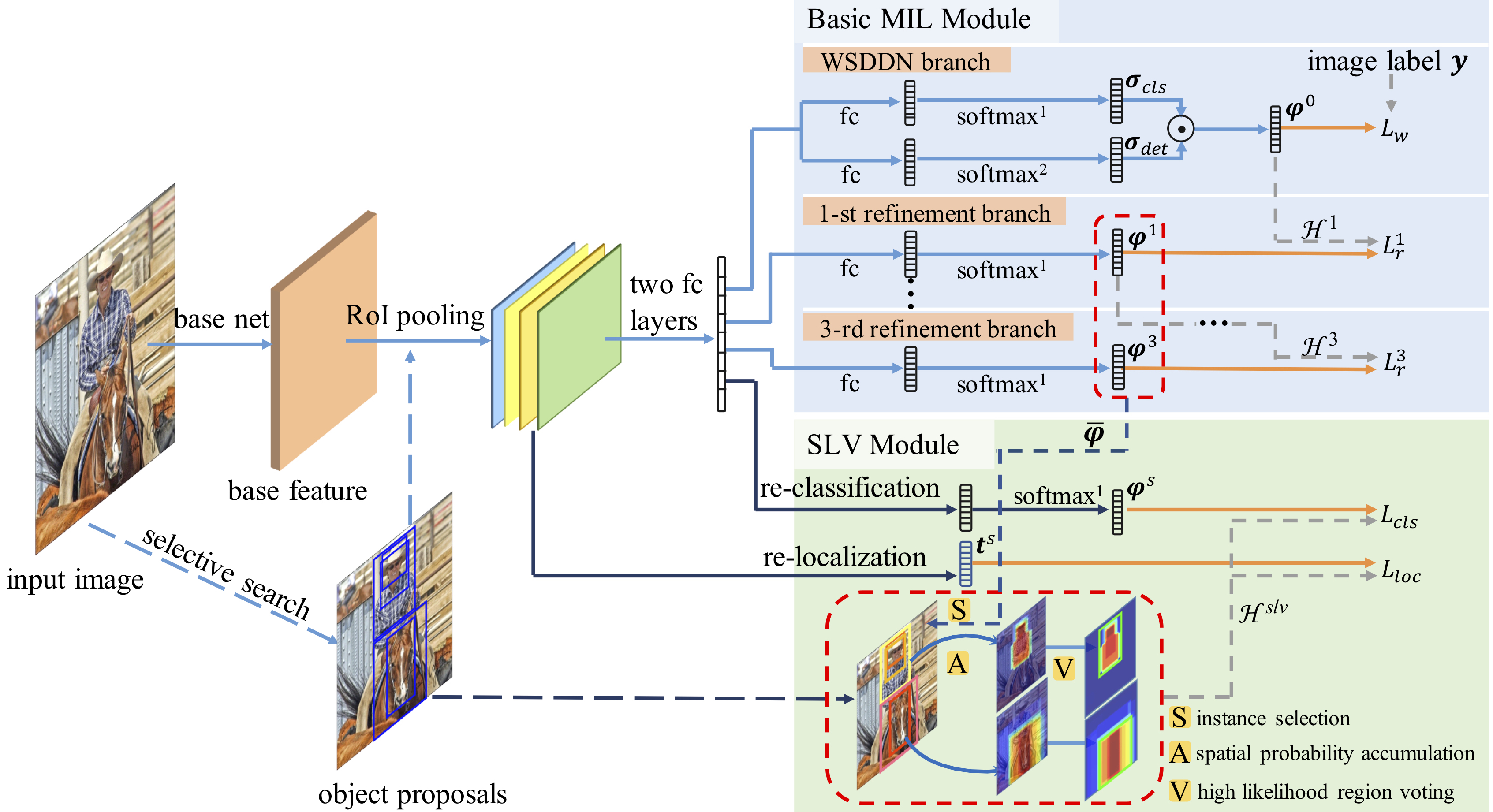}
\end{center}
   \caption{The network architecture of our method. A VGG16 base net with RoI pooling is used to extract the
   feature of each proposal. Then the proposal features pass through two fully connected layers and the generated
   feature vectors are branched into Basic MIL module and SLV module (re-classification branch). In Basic MIL
   Module, there are one WSDDN branch and three refinement branches. The average classification scores of three
   refinement branches are fed into SLV module to generate supervisions. Another fully connected layer in SLV
   module is used to obtain regression offsets (re-localization branch). $softmax^1$ is softmax operation over
   classes and $softmax^2$ is softmax operation over proposals.
    }
\label{fig:architecture}
\end{figure*}

Besides, there are many attempts~\cite{arun2019dissimilarity,kosugi2019object,li2019weakly,
yang2019towards,zhang2018w2f} to improve the localization accuracy of the weakly supervised detectors from other
perspectives.
Arun \etal~\cite{arun2019dissimilarity} obtain much better performance by employing a probabilistic
objective to model the uncertainty in the location of objects.
In~\cite{li2019weakly}, Li \etal propose a segmentation-detection collaborative network which utilizes
the segmentation maps as prior information to supervise the learning of object detection. 
In~\cite{kosugi2019object}, Kosugi \etal focus on instance labeling problem and design two different labeling methods to
find tight boxes rather than discriminative ones.
In~\cite{zhang2018w2f}, Zhang \etal propose to mine accurate pseudo ground-truths from a well-trained MIL-based network to train
a fully supervised object detector.
In contrast, the work of Yang \etal~\cite{yang2019towards} integrates WSOD and Fast-RCNN re-training
into a single network that can jointly optimize the regression and classification.


\section{Method}
The overall architecture of the proposed framework is shown
in Fig \ref{fig:architecture}. We adopt a MIL-based network as a basic part and integrate the proposed SLV module
into the final architecture. During the forward process of training, the proposal features are fed into the basic MIL
module to produce proposal score matrices. Subsequently, these proposal score matrices are used to generate
supervisions for the training of the proposed SLV module.

\subsection{Basic MIL Module} \label{section3-1}

With image-level annotations, many existing works~\cite{bilen2015weakly,bilen2016weakly,cinbis2016weakly,kantorov2016contextlocnet}
detect objects based on a MIL network. In this work, we follow the method in~\cite{bilen2016weakly} which
proposes a two-stream weakly supervised deep detection network (WSDDN) to train the instance classifier. For a
training image and its region proposals, the proposal features are extracted by a CNN backbone and then branched
into two streams, which correspond to a classification branch and a detection branch respectively. For classification
branch, the score matrix ${\boldsymbol{X}}^{cls}$ $\in$ $\mathbb{R}^{C\times R}$ is produced by passing the proposal
features through a fully connected (fc) layer, where $C$ denotes the number of image classes and $R$ denotes the number
of proposals. Then a softmax operation over classes is performed to produce $\boldsymbol{\sigma}_{cls} ( {\boldsymbol {X}}^{cls} )$,
$[\sigma_{cls} ( {\boldsymbol {X}}^{cls} )]_{cr} = \frac{e^{{x_{cr}^{cls}}}} { \sum_{k=1}^{C}e^{x_{kr}^{cls}}}$.
Similarly, the score matrix ${\boldsymbol{X}}^{det}$
$\in$ $\mathbb{R}^{C\times R}$ is produced by another fc layer for detection branch, but
$\boldsymbol{\sigma}_{det}({\boldsymbol {X}}^{det})$ is generated through a softmax operation over proposals rather than classes:
$[ \sigma_{det}( {\boldsymbol {X}}^{det} ) ]_{cr} = \frac{e^{{x_{cr}^{det}}}} { \sum_{k=1}^{ R }e^{x_{ck}^{det}}}$.
The score of each proposal is generated by element-wise product:
$\boldsymbol{\varphi}^0 = \boldsymbol{\sigma}_{cls} ( {\boldsymbol {X}}^{cls} )\odot \boldsymbol{\sigma}_{det} ( {\boldsymbol {X}}^{det} ) $.
At last, the image classification score on class $c$ is computed through the summation over all proposals: 
$\phi_c = \sum_{r=1}^{R}\varphi_{cr}^0$.
We denote the label of a training image ${\boldsymbol {y} } = \left [ y_1,y_2,...,y_C \right ]^{T}$,
where $y_c = 1$ or 0 indicates the image with or without class $c$. To train the instance classifier, the loss function is shown in Eq. (\ref{loss_w}).
\begin{equation}
L_w = -\sum_{c=1}^{C}\left \{ y_c\log \phi_c + (1 - y_c)\log (1 - \phi_c) \right \} \label{loss_w}
\end{equation}

Moreover, proposal cluster learning (PCL)~\cite{tang2018pcl} is adopted, which embeds 3
instance classifier refinement branches additionally, to get better instance classifiers. 
The output of the $k$-th refinement branch is $\boldsymbol {\varphi}^k \in \mathbb{R}^{(C + 1)\times R}$,
where $(C + 1)$ denotes the number of $C$ different classes and background.


Specifically, based on the output score $\boldsymbol {\varphi}^k$ and proposal spatial information,
proposal cluster centers are built.
All proposals are then divided into those clusters according to the $IoU$ between them, one for background and the others for different instances.
Proposals in the same cluster (except for the cluster for background) are spatially adjacent and associated with the same object.
With the supervision $\mathcal{H}^k = \left \{ y_n^k \right \}_{n=1}^{N^k+1}$ ($y_n^k$ is the label of the 
$n$-th cluster), the refinement branch treats each cluster as a small bag. Each bag in the $k$-th refinement
branch is optimized by a weighted cross-entropy loss.

\begin{equation}
\begin{aligned}
L^k = -\frac{1}{R}&( \sum_{n=1}^{N^k} s_n^k M_n^k \log \frac{\sum\limits_{r \in \mathcal{C}_n^k} \varphi_{y_n^kr}^k }{M_n^k} \\
&+ \sum\limits_{r \in \mathcal{C}_{N^k+1}^k} \lambda_r^k \log \varphi_{(C+1)r}^k) \label{loss_k}
\end{aligned}
\end{equation}
where $s_n^k$ and $M_n^k$ are the confidence score of $n$-th cluster and the number of proposals in the $n$-th cluster,
$\varphi_{cr}^k$ is the predicted score of the $r$-th proposal.
$r \in \mathcal{C}_n^k$ indicates that the $r$-th proposal belongs to the $n$-th
proposal cluster, $\mathcal{C}_{N^k+1}^k$ is the cluster for background,
$\lambda_r^k$ is the loss weight that is the same as the confidence of the $r$-th proposal. 


\subsection{Spatial Likelihood Voting} \label{section 3-2}

It is hard for weakly supervised object detectors to pick out the most appropriate bounding boxes from all proposals for an object.
The proposal that obtains the highest classification score often covers a discriminative part of an object while many other proposals covering the larger parts tend to have lower scores.
Therefore, it is unstable to choose the proposal with the highest score as the detection result under the MIL constraints.
But from the overall distribution, those high-scoring proposals always cover at least parts of objects.
To this end, we propose to make use of the spatial likelihood of all proposals which implies the boundaries and categories of objects in an image.
In this subsection, we introduce a spatial likelihood voting (SLV) module to perform classification and localization refinement simultaneously rather than the instance classifier only.

The SLV module is convenient to be plugged into any proposal-based detector and can be optimized with the fundamental detector jointly.
The spirit of SLV is to establish a bridge between classification task and localization task through coupling the spatial information and category information of all proposals together.
During training, the SLV module takes into the classification scores of all proposals and then calculates the spatial likelihood of them for generating supervision $\mathcal{H}^{slv}\left ( \bar{\boldsymbol{\varphi}}, \boldsymbol{{y}}\right )$, where $\bar{\boldsymbol{\varphi}} = \left ( \sum_{k=1}^{3} \boldsymbol{\varphi}^k \right )/3$.



Formally, for an image $\boldsymbol{\rm{I}}$ with label $\boldsymbol{y}$, there are three steps to generate $\mathcal{H}^{slv}_c$ when $y_c = 1$.
To save training time, the low-scoring proposals are filtered out first as they have little significance for spatial likelihood voting.
The retained proposals are considered to surround the instances of category $c$ and are placed into  $\mathcal{B}_c$, $\mathcal{B}_c = \{b_{r}\ |\ \bar{\varphi}_{cr} > T_{score}\}$.

For the second step, we implement a spatial probability accumulation according to the predicted classification
scores and locations of proposals in $\mathcal{B}_c$.
In detail, we construct a score matrix $\boldsymbol{{M}}^c$ $\in$ $\mathbb{R}^{H\times W}$, where $H$ and $W$
are the height and width of the training image $\boldsymbol{\rm{I}}$.
All elements in $\boldsymbol{{M}}^c$ are initialized with zero. Then, for each proposal $b_r \in \mathcal{B}_c$,
we accumulate the predicted score of $b_r$ to $\boldsymbol{{M}}^c$ spatially.
\begin{equation}
{\rm{m}}_{ij}^c=\sum_{\substack{r\;s.t.\;b_r \in \mathcal{B}_c, \\ \left (i,j\right) \in b_r }} \bar{\varphi}_{cr} \label{Mc}
\end{equation}
where $\left(i,j\right) \in b_r$ means the pixel $\left (i,j\right)$ inside the proposal $b_r$.
For proposals in $\mathcal{B}_c$, we calculate their likelihood in spatial dimensions and the final value
of elements in $\boldsymbol{{M}}^c$ indicates the possibility that the instance of category $c$ appears in that position.

\begin{algorithm}[tb] 
\caption{Generating supervision $\mathcal{H}^{slv}$} 
\label{alg:algo1} 
\begin{algorithmic}[1] 
\REQUIRE ~~\\ 
Proposal boxes $\mathcal{B} = \left \{b_1,...,b_R\right \}$; proposal average scores $\bar{\boldsymbol{\varphi}}$;
image label vector $\boldsymbol{{y}} = \left [y_1,...,y_C\right ]^T$; image size $\left \{H, W\right \}$.
\ENSURE ~~\\ 
Supervision $\mathcal{H}^{slv}$.
\STATE Initialize $\mathcal{H}^{slv} = \varnothing$.\\
\FOR{$c = 1$ to $C$}
\IF{$y_c = 1$}
\STATE Initialize $\mathcal{B}_c = \varnothing$.\\
Initialize $\boldsymbol{{M}}^c$ with zero.
\FOR{$r = 1$ to $R$}
\IF{$\bar{{\varphi}}_{cr} > T_{score}$}
\STATE $\mathcal{B}_c$.append($b_r$).
\ENDIF
\ENDFOR
\STATE Construct $\boldsymbol{M}^c$ by Eq. (\ref{Mc}), see Section \ref{section 3-2}.
\STATE Scale the range of elements in $\boldsymbol{{M}}^c$ to $\left [ 0, 1 \right ]$.
\STATE Transform $\boldsymbol{M}^c$ into the binary version $\boldsymbol{M}_b^c$.
\STATE Find minimum bounding rectangles $\mathcal{G}_c$ in $\boldsymbol{M}_b^c$.
\STATE $\mathcal{H}_c^{slv} = \left \{ \mathcal{G}_c, c \right \}$.
\STATE $\mathcal{H}^{slv}$.append($\mathcal{H}_c^{slv}$).
\ENDIF
\ENDFOR
\RETURN $\mathcal{H}^{slv}$. 
\end{algorithmic}
\end{algorithm}

Finally, the range of elements in $\boldsymbol{{M}}^c$ is scaled to $[0, 1]$ and a threshold $T_b^c$ is set to transform $\boldsymbol{{M}}^c$ into a binary version $\boldsymbol{{M}}_b^c$.
$\boldsymbol{{M}}_b^c$ is regarded as a binary image and the minimum bounding
rectangles $\mathcal{G}_c =\left \{ g_m \right \}_{m=1}^{N_c}$ of connected regions 
in $\boldsymbol{{M}}_b^c$ ($g_m$ is the $m$-th rectangle and $N_c$ is the number of connected regions in $\boldsymbol{{M}}_b^c$)
is used to generate $\mathcal{H}_c^{slv}$ shown in Eq. (\ref{H_c_slv}).
\begin{equation}
\mathcal{H}_c^{slv} = \left \{ \mathcal{G}_c, c \right \} \label{H_c_slv}
\end{equation}

\begin{algorithm}[tb] 
\caption{The overall training procedure} 
\label{alg:algo2} 
\begin{algorithmic}[1] 
\REQUIRE ~~\\ 
A training image and its proposal boxes $\mathcal{B}$; 
image label vector $\boldsymbol{{y}} = \left [y_1,...,y_C\right ]^T$;
refinement times $K=3$;
training iteration index $i$.
\ENSURE ~~\\ 
An updated network.
\STATE Feed the image and proposal boxes $\mathcal{B}$ into basic MIL module
to produce score matrices $\boldsymbol {\varphi}^k,k \in \left \{0, 1, 2, 3\right \}$
\STATE Compute loss $L_w$ and $L_r^k$, $k \in \left\{1,2,3\right\}$ by Eq. (\ref{loss_k})/(\ref{loss_w}), see Section \ref{section3-1}.
\STATE Compute average score matrix $\bar{\boldsymbol{\varphi}} = \left ( \sum_{k=1}^{3} \boldsymbol{\varphi}^k \right )/3$.
\FOR{$c = 1$ to $C$}
\IF{$y_c = 1$}
\STATE Generate $\mathcal{H}_c^{slv}$ based on $\bar{\boldsymbol{\varphi}}$ and proposal boxes $\mathcal{B}$, see Section \ref{section 3-2}.
\ENDIF
\ENDFOR
\STATE Generate $\mathcal{H}^{slv}$, see Algorithm \ref{alg:algo1}.
\STATE Compute loss $L_{s}$, see Section \ref{section 3-2}.
\STATE Compute loss weight $w_{s}$ by training iteration index $i$.
\STATE Optimize ($L_w + \sum_{k=1}^{3}L_r^k + w_{s}L_{s}$).
\end{algorithmic}
\end{algorithm}

%

\begin{figure}[t]
\begin{center}
\includegraphics[width=1.0\linewidth]{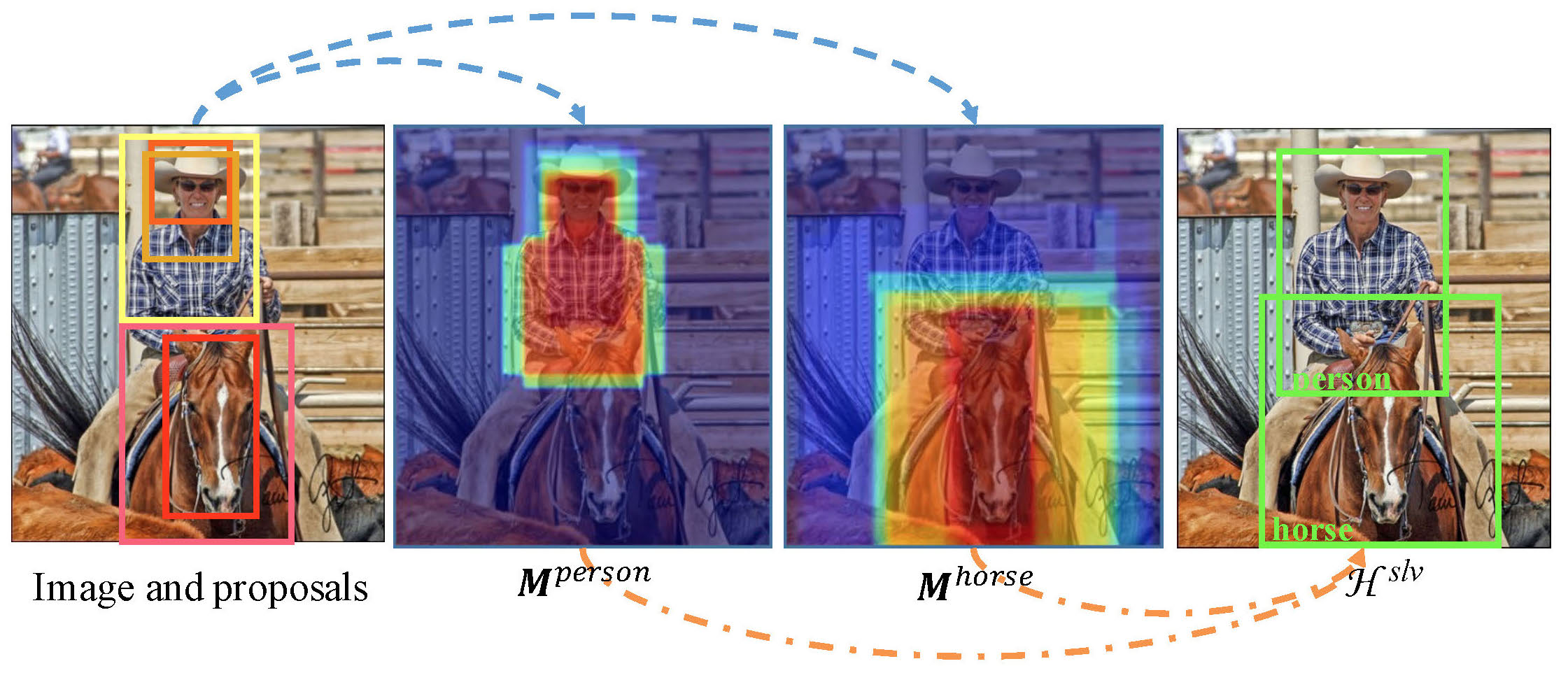}
\end{center}
   \caption{A visualization example of SLV. The label of image is $\left \{person, horse\right \}$,
   then two different $\boldsymbol{M}^c$ and $\mathcal{H}^{slv}$ are generated.}
\label{fig:slv present}
\end{figure}

The overall procedures of generating $\mathcal{H}^{slv}$ are summarized in Algorithm \ref{alg:algo1} and a
visualization example of SLV is shown in Fig \ref{fig:slv present}.
Supervision $\mathcal{H}^{slv}$ is instance-level annotation and we use a multi-task loss $L_{s}$ on each
labeled proposal to perform classification and localization refinement simultaneously.
The output of re-classification branch is $\boldsymbol{\varphi}^{s} \in \mathbb{R}^{(C + 1)\times R}$ and
output of re-localization branch is $\boldsymbol{t}^{s} \in \mathbb{R}^{4\times R}$.
The loss of SLV module is $L_{s} = L_{cls}(\boldsymbol{\varphi}^{s}, \mathcal{H}^{slv}) + L_{loc}(\boldsymbol{t}^{s}, \mathcal{H}^{slv})$,
where $L_{cls}$ is the cross entropy loss and $L_{loc}$ is the smooth L1 loss.

\subsection{The overall training framework} \label{section 3-3}
To refine the weakly supervised object detector, the basic MIL module and SLV module are integrated into one.
Combining the loss function of both, the final loss of the whole network is in Eq. (\ref{loss_all}).
\begin{equation}
L = L_w + \sum\nolimits_{k=1}^{3}L_r^k + L_{s} \label{loss_all}
\end{equation}

However, the output classification scores of basic MIL module are noisy in early stage of training,
which causes that the voted supervisions $\mathcal{H}^{slv}$ are not precise enough to train the object detector.
There is an alternative training strategy to avoid this problem: 1) fixing the SLV module and training the basic MIL module completely;
2) fixing the basic MIL module and using the output classification scores of it to train the
SLV module. This strategy makes sense but training different parts of the network separately may harm
the performance. So, we propose a training framework that integrates the two training steps into one. We change the loss
in Eq. (\ref{loss_all}) to a weighted version, as in Eq. (\ref{loss_final}).
\begin{equation}
L = L_w + \sum\nolimits_{k=1}^{3}L_r^k + w_{s}L_{s} \label{loss_final}
\end{equation}
The loss weight $w_{s}$ is initialized with zero and will increase iteratively. At the beginning of training, although the
basic MIL module is unstable and we cannot obtain good supervisions $\mathcal{H}^{slv}$, $w_{s}$ is small and
the loss $w_{s}L_{s}$ is also small. As a consequence, the performance of the basic MIL module will
not be affected a lot. During the training process, the basic MIL module will classify the proposals well, and
thus we can obtain stable classification scores to generate more precise supervisions $\mathcal{H}^{slv}$. 
The proposed training framework is easy to implement and the network could benefit from the shared proposal features. The overall
training procedure of our network is shown in Algorithm \ref{alg:algo2}. 

During testing, the proposal scores of three refined instance classifiers and SLV re-classification branch are used
as the final detection scores. And the bounding box regression offsets computed by the SLV re-localization branch are
used to shift all proposals.


\section{Experiment}

\subsection{Datasets and Evaluation Metrics}
SLV was evaluated on two challenging datasets: PASCAL VOC 2007 and 2012 datasets~\cite{everingham2015pascal}
which have 9,962 and 22,531 images respectively for 20 object classes.
For each dataset, we use the $trainval$ set for training and $test$ set for testing. Only image-level
annotations are used to train our network.

\begin{table}[t]
\begin{center}
\begin{tabular}{c c c c|c}
\hline
re-cls &re-loc&end-to-end&fast-rcnn&mAP\\
\hline\hline
 & & & &50.1  \\
\checkmark& & & &51.0  \\
 &\checkmark& & &51.6 \\
\checkmark&\checkmark& & &52.5  \\
\checkmark&\checkmark&\checkmark&&53.5 \\
\checkmark&\checkmark&\checkmark&\checkmark&\bfseries{53.9} \\
\hline
\end{tabular}
\end{center}
\caption{Detection performance for different ablation experiments on PASCAL VOC 2007 $test$ set.
``$re$-$cls$'' and ``$re$-$loc$'' means re-classification and re-localization branch respectively.
``$end$-$to$-$end$'' is the proposed training framework and ``$fast$-$rcnn$'' means re-training a Fast-RCNN detector.}
\label{ablation-compare}
\end{table}

For evaluation, two metrics are used to evaluate our model. First, we evaluate detection performance using
mean Average Precision (mAP) on the PASCAL VOC 2007 and 2012 $test$ set.
Second, we evaluate the localization accuracy using Correct Localization (CorLoc) on PASCAL VOC 2007 and 2012 $trainval$ set.
Based on the PASCAL criterion, a predicted box is considered positive if it has an $IoU > 0.5$ with a ground-truth bounding box.

\subsection{Implementation Details}
The proposed framework is implemented based on VGG16~\cite{simonyan2014very} CNN model, which is pre-trained on ImageNet dataset~\cite{russakovsky2015imagenet}.
We use Selective Search~\cite{uijlings2013selective} to generate about 2,000 proposals per-image.
In basic MIL module, we follow the implementation in~\cite{tang2018pcl} to refine instance classifier three times.
For SLV module, we use the average proposal scores of three refined instance classifiers to generate supervisions and the setting of hyper-parameters is intuitive. The threshold $T_{score}$ is set to 0.001 for saving time and $T_b^c$ is set to 0.2 for person category and 0.5 for other categories.

During training, the mini-batch size for training is set to 2. The momentum and weight decay are set to 0.9
and $5 \times 10^{-4}$ respectively. The initial learning rate is $5 \times 10^{-4}$ and the learning rate decay step is 9-th, 12-th and 15-th epoch.
For data augmentation, we use five image scales $\left \{ 480, 576, 688, 864, 1200\right \}$
with horizontal flips for both training and testing. We randomly choose a scale to resize the image and then the image is
horizontal flipped. During testing, the average score of 10 augmented images is used as the final classification score. Similarly, 
the output regression offsets of 10 augmented images are also averaged.

Our experiments are implemented based on PyTorch\cite{paszke2017automatic} deep learning framework.
And all of our experiments are running on NVIDIA GTX 1080Ti GPU.

\subsection{Ablation Studies}

We perform ablations on PASCAL VOC 2007 to analyze the proposed SLV module.
The baseline model(mAP 50.1\% on PASCAL VOC 2007 $test$ set) is the basic PCL
detector described in Section \ref{section3-1}, which is trained on PASCAL
VOC 2007 $trainval$ set. 
Details about ablation studies are discussed in the following.

\begin{figure}[t]
\begin{center}
   \includegraphics[width=1.0\linewidth]{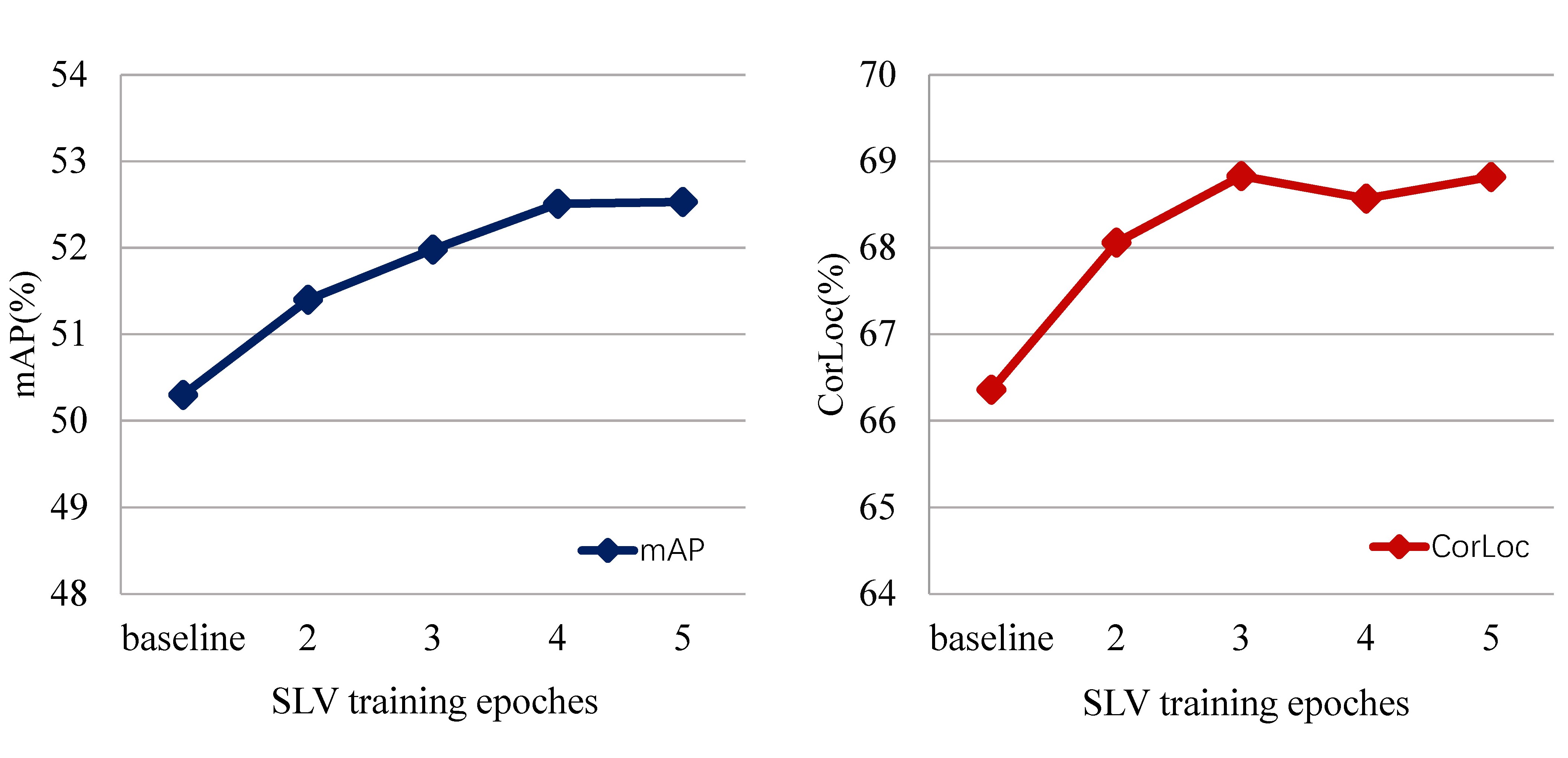}
\end{center}
   \caption{Results on VOC 2007 for baseline and different 
   training epochs of SLV module.}
\label{fig:ablation1}
\end{figure}

\begin{figure}[t]
\begin{center}
   \includegraphics[width=1\linewidth]{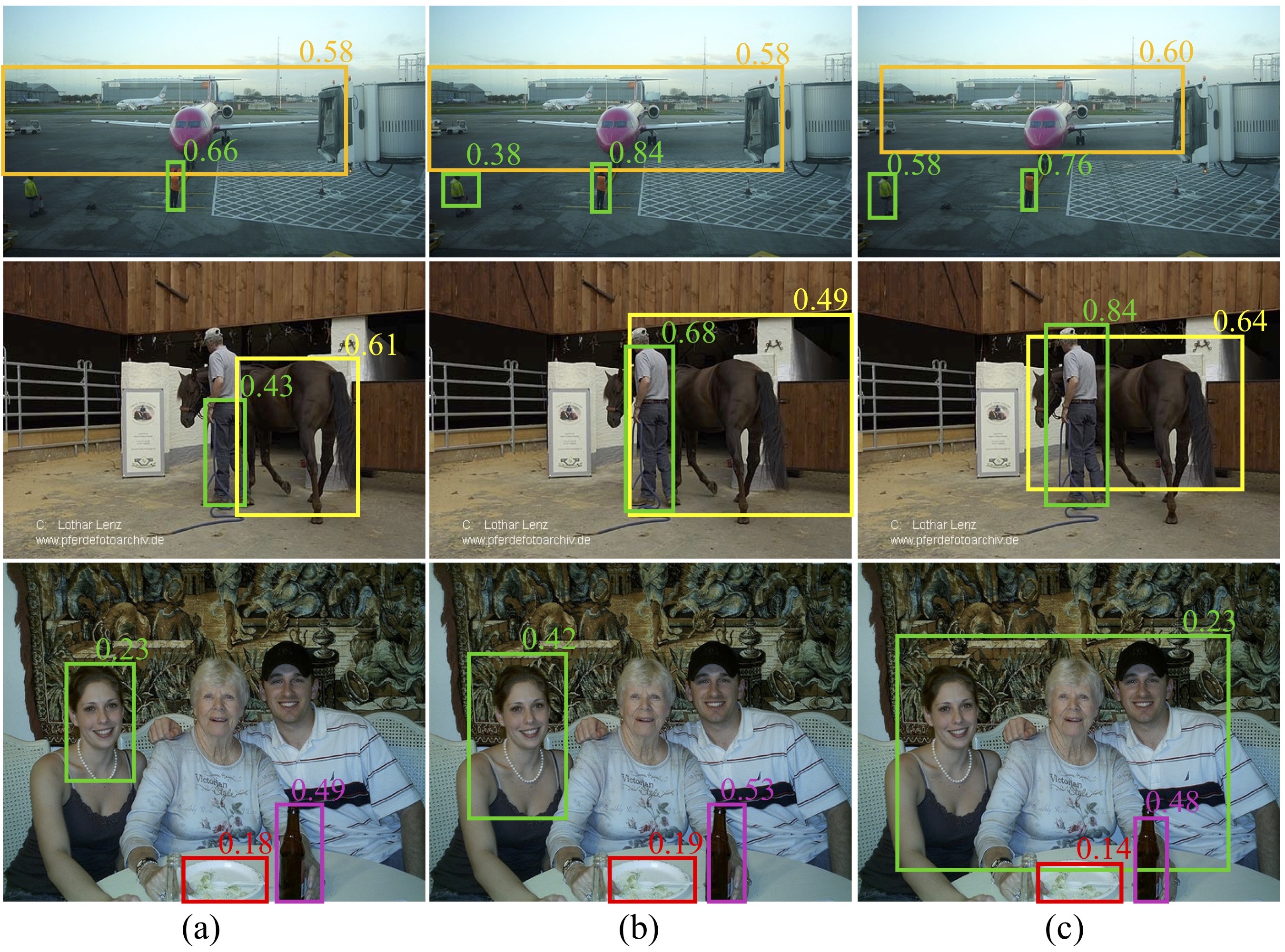}
\end{center}
   \caption{Examples of 3 different labeling schemes.
      (a) Conventional scheme.
      (b) Clustering scheme.
      (c) SLV.
      The value on the top of every labeled box is the $IoU$ with its corresponding ground-truth bounding box.
}
\label{fig:label}
\end{figure}

\textbf{SLV} \textit{vs.} \textbf{No SLV.}
To confirm the effectiveness of the proposed SLV module, we conduct different
ablation experiments for re-classification and re-localization branch in SLV.
As shown in Table \ref{ablation-compare} (row 2 and row 3), the simplified versions of SLV module which
only contain a re-classification or re-localization branch both outperform the baseline model. It 
indicates the supervision generated by spatial likelihood voting method, which is formulated in Section \ref{section 3-2},
is precise enough not only for classification but also for localization.

Moreover, a normal version of SLV module improves the detection performance further due to multi-task learning.
As shown in Fig \ref{fig:ablation1}, the SLV module trained based on a well-trained baseline model boosts the
performance significantly (mAP from 50.1\% to 52.5\%), indicating the 
necessity of converging the proposal localizing process into WSOD solutions as we discussed above.

\begin{table*}[htbp]
\begin{center}
\resizebox{\textwidth}{!}{
\begin{tabular}{l|cccccccccccccccccccc|c}
\hline
Method &aero&bike&bird&boat&bottle&bus&car&cat&chair&cow&table&dog&horse&mbike&person&plant&sheep&sofa&train&tv&mAP \\
\hline\hline
OICR(VGG)~\cite{tang2017multiple}    &58.0&62.4&31.1&19.4&13.0&65.1&62.2&28.4&24.8&44.7&30.6&25.3&37.8&65.5&15.7&24.1&41.7&46.9&64.3&62.6&41.2 \\
PCL(VGG)~\cite{tang2018pcl}          &54.4&69.0&39.3&19.2&15.7&62.9&64.4&30.0&25.1&52.5&44.4&19.6&39.3&67.7&17.8&22.9&46.6&57.5&58.6&63.0&43.5 \\
WS-RPN(VGG)~\cite{tang2018weakly}    &57.9&70.5&37.8&5.7&21.0&66.1&69.2&59.4&3.4&57.1&\bfseries{57.3}&35.2&64.2&68.6&\bfseries{32.8}&28.6&50.8&49.5&41.1&30.0&45.3 \\
C-MIL~\cite{wan2019c}                 &62.5&58.4&49.5&32.1&19.8&70.5&66.1&63.4&20.0&60.5&52.9&53.5&57.4&68.9&8.4&24.6&51.8&58.7&66.7&63.5&50.5 \\
UI~\cite{gao2019utilizing}            &63.4&70.5&45.1&28.3&18.4&69.8&65.8&69.6&27.2&62.6&44.0&59.6&56.2&\bfseries{71.4}&11.9&26.2&\bfseries{56.6}&59.6&\bfseries{69.2}&\bfseries{65.4}&52.0 \\
Pred Net(VGG)~\cite{arun2019dissimilarity} &\bfseries{66.7}&69.5&\bfseries{52.8}&31.4&\bfseries{24.7}&\bfseries{74.5}&\bfseries{74.1}&67.3&14.6&53.0&46.1&52.9&\bfseries{69.9}&70.8&18.5&28.4&54.6&\bfseries{60.7}&67.1&60.4&52.9 \\
SLV(VGG)                              &65.6&\bfseries{71.4}&49.0&\bfseries{37.1}&24.6&69.6&70.3&\bfseries{70.6}&\bfseries{30.8}&\bfseries{63.1}&36.0&\bfseries{61.4}&65.3&68.4&12.4&\bfseries{29.9}&52.4&60.0&67.6&64.5&\bfseries{53.5}\\
\hline\hline
OICR+FRCNN~\cite{tang2017multiple} &65.5&67.2&47.2&21.6&22.1&68.0&68.5&35.9&5.7&63.1&49.5&30.3&64.7&66.1&13.0&25.6&50.0&57.1&60.2&59.0&47.0 \\
PCL+FRCNN~\cite{tang2018pcl}       &63.2&69.9&47.9&22.6&27.3&71.0&69.1&49.6&12.0&60.1&51.5&37.3&63.3&63.9&15.8&23.6&48.8&55.3&61.2&62.1&48.8 \\
WS-RPN+FRCNN~\cite{tang2018weakly} &63.0&69.7&40.8&11.6&\bfseries{27.7}&70.5&\bfseries{74.1}&58.5&10.0&\bfseries{66.7}&\bfseries{60.6}&34.7&\bfseries{75.7}&\bfseries{70.3}&25.7&26.5&\bfseries{55.4}&56.4&55.5&54.9&50.4 \\
W2F~\cite{zhang2018w2f}            &63.5&70.1&50.5&31.9&14.4&72.0&67.8&73.7&23.3&53.4&49.4&65.9&57.2&67.2&\bfseries{27.6}&23.8&51.8&58.7&64.0&62.3&52.4 \\
UI+FRCNN~\cite{gao2019utilizing}   &62.7&69.1&43.6&31.1&20.8&69.8&68.1&72.7&23.1&65.2&46.5&64.0&67.2&66.5&10.7&23.8&55.0&\bfseries{62.4}&\bfseries{69.6}&60.3&52.6 \\
C-MIL+FRCNN~\cite{wan2019c}        &61.8&60.9&\bfseries{56.2}&28.9&18.9&68.2&69.6&71.4&18.5&64.3&57.2&66.9&65.9&65.7&13.8&22.9&54.1&61.9&68.2&\bfseries{66.1}&53.1 \\
Pred Net(Ens)~\cite{arun2019dissimilarity} &\bfseries{67.7}&70.4&52.9&31.3&26.1&\bfseries{75.5}&73.7&68.6&14.9&54.0&47.3&53.7&70.8&70.2&19.7&\bfseries{29.2}&54.9&61.3&67.6&61.2&53.6 \\
SLV(VGG)+FRCNN                      &62.1&\bfseries{72.1}&54.1&\bfseries{34.5}&25.6&66.7&67.4&\bfseries{77.2}&\bfseries{24.2}&61.6&47.5&\bfseries{71.6}&72.0&67.2&12.1&24.6&51.7&61.1&65.3&60.1&\bfseries{53.9}\\
\hline
\end{tabular}}
\end{center}
\caption{Average precision (in $\%$) on PASCAL VOC 2007 test set. The first part shows the results of weakly supervised object detectors
using a single model and the second part shows the results of weakly supervised object detectors using an ensemble model or fully supervised
object detector trained by pseudo ground-truths generated by weakly supervised object detectors.}
\label{mAP-2007}
\end{table*}

\begin{table*}[htbp]
\begin{center}
\resizebox{\textwidth}{!}{
\begin{tabular}{l|cccccccccccccccccccc|c}
\hline
Method &aero&bike&bird&boat&bottle&bus&car&cat&chair&cow&table&dog&horse&mbike&person&plant&sheep&sofa&train&tv&CorLoc \\
\hline\hline
OICR(VGG)~\cite{tang2017multiple} &81.7&80.4&48.7&49.5&32.8&81.7&85.4&40.1&40.6&79.5&35.7&33.7&60.5&88.8&21.8&57.9&76.3&59.9&75.3&81.4&60.6 \\
PCL(VGG)~\cite{tang2018pcl}       &79.6&85.5&62.2&47.9&37.0&83.8&83.4&43.0&38.3&80.1&50.6&30.9&57.8&90.8&27.0&58.2&75.3&68.5&75.7&78.9&62.7 \\
WS-RPN(VGG)~\cite{tang2018weakly} &77.5&81.2&55.3&19.7&44.3&80.2&86.6&69.5&10.1&\bfseries{87.7}&\bfseries{68.4}&52.1&84.4&91.6&\bfseries{57.4}&\bfseries{63.4}&77.3&58.1&57.0&53.8&63.8 \\

C-MIL~\cite{wan2019c}              &-&-&-&-&-&-&-&-&-&-&-&-&-&-&-&-&-&-&-&-&65.0 \\
UI~\cite{gao2019utilizing}         &84.2&84.7&59.5&52.7&37.8&81.2&83.3&72.4&41.6&84.9&43.7&69.5&75.9&90.8&18.1&54.9&81.4&60.8&79.1&80.6&66.9 \\
Pred Net(VGG)~\cite{arun2019dissimilarity} &\bfseries{88.6}&\bfseries{86.3}&71.8&53.4&\bfseries{51.2}&\bfseries{87.6}&\bfseries{89.0}&65.3&33.2&86.6&58.8&65.9&87.7&\bfseries{93.3}&30.9&58.9&\bfseries{83.4}&\bfseries{67.8}&78.7&80.2&70.9 \\
SLV(VGG)                           &84.6&84.3&\bfseries{73.3}&\bfseries{58.5}&49.2&80.2&87.0&\bfseries{79.4}&\bfseries{46.8}&83.6&41.8&\bfseries{79.3}&\bfseries{88.8}&90.4&19.5&59.7&79.4&67.7&\bfseries{82.9}&\bfseries{83.2}&\bfseries{71.0}\\
\hline\hline
OICR+FRCNN~\cite{tang2017multiple} &85.8&82.7&62.8&45.2&43.5&84.8&87.0&46.8&15.7&82.2&51.0&45.6&83.7&91.2&22.2&59.7&75.3&65.1&76.8&78.1&64.3 \\
PCL+FRCNN~\cite{tang2018pcl}       &83.8&85.1&65.5&43.1&50.8&83.2&85.3&59.3&28.5&82.2&57.4&50.7&85.0&92.0&27.9&54.2&72.2&65.9&77.6&\bfseries{82.1}&66.6 \\
WS-RPN+FRCNN~\cite{tang2018weakly} &83.8&82.7&60.7&35.1&\bfseries{53.8}&82.7&88.6&67.4&22.0&86.3&\bfseries{68.8}&50.9&\bfseries{90.8}&93.6&\bfseries{44.0}&\bfseries{61.2}&82.5&65.9&71.1&76.7&68.4 \\
W2F~\cite{zhang2018w2f}            &-&-&-&-&-&-&-&-&-&-&-&-&-&-&-&-&-&-&-&-&70.3 \\
UI+FRCNN~\cite{gao2019utilizing}   &86.7&85.9&64.3&55.3&42.0&84.8&85.2&78.2&47.2&\bfseries{88.4}&49.0&73.3&84.0&92.8&20.5&56.8&\bfseries{84.5}&62.9&\bfseries{82.1}&78.1&66.9 \\
Pred Net(Ens)~\cite{arun2019dissimilarity} &\bfseries{89.2}&\bfseries{86.7}&72.2&50.9&51.8&\bfseries{88.3}&\bfseries{89.5}&65.6&33.6&87.4&59.7&66.4&88.5&\bfseries{94.6}&30.4&60.2&83.8&68.9&78.9&81.3&71.4 \\
SLV(VGG)+FRCNN                     &85.8&85.9&\bfseries{73.3}&\bfseries{56.9}&52.7&79.7&87.1&\bfseries{84.0}&\bfseries{49.3}&82.9&46.8&\bfseries{81.2}&89.8&92.4&21.2&59.3&80.4&\bfseries{70.4}&\bfseries{82.1}&78.8&\bfseries{72.0}\\

\hline
\end{tabular}}
\end{center}
\caption{CorLoc (in $\%$) on PASCAL VOC 2007 trainval set. 
The first part shows the results of weakly supervised object detectors
using a single model and the second part shows the results of weakly supervised object detectors
using an ensemble model or fully supervised object detector trained by pseudo ground-truths generated by weakly supervised object detectors.}
\label{corloc-2007}
\end{table*}

\begin{table}[t]
\begin{center}
\begin{tabular}{l|c c}
\hline
Method &mAP(\%)&CorLoc(\%)\\
\hline\hline
PCL(VGG)~\cite{tang2018pcl}          &40.6&63.2 \\
WS-RPN(VGG)~\cite{tang2018weakly}    &40.8&64.9 \\
C-MIL~\cite{wan2019c}                 &46.7&67.4 \\
UI~\cite{gao2019utilizing}            &48.0&67.4 \\
Pred Net(VGG)~\cite{arun2019dissimilarity} &48.4&\bfseries{69.5} \\
SLV(VGG)                              &\bfseries{49.2}&69.2 \\
\hline
\end{tabular}
\end{center}
\caption{Detection and localization performance for different detectors using a single model on PASCAL VOC 2012 dataset.}
\label{results-2012}
\end{table}

\textbf{End-to-end} \textit{vs.} \textbf{Alternative.}
In the previous subsection, the ablation experiments are conducted by the way that fixes the
baseline model and trains the SLV module only. 
The two parts of the proposed network are trained separately, which
is similar to re-train an independent Fast-RCNN model.

In the row 4 and 5 of Table \ref{ablation-compare}, we present the performance of models with different training strategies. 
Compared with the alternative training
strategy (row 4), the model trained with the proposed end-to-end training framework (row 5)
outperforms the former a lot. Just as we discussed in Section
\ref{section 3-3}, end-to-end training framework shorten the gap between weakly-supervised and fully-supervised object
detection.

\begin{figure*}
\begin{center}
\includegraphics[width=6.0in]{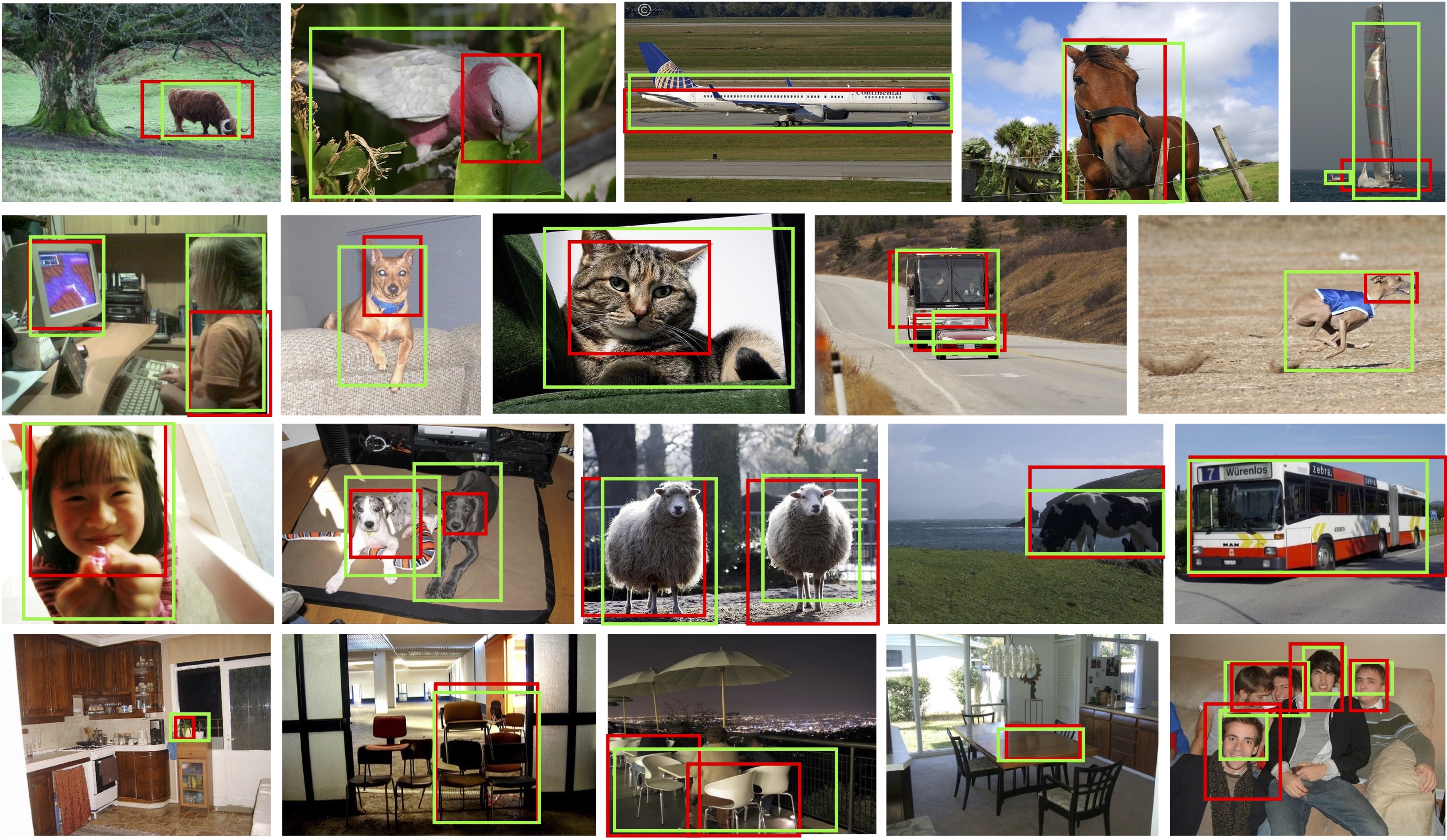}
\end{center}
   \caption{Detection results of our method and a competitor (the PCL model). Green
   bounding boxes are the objects detected by our method and red ones are the results detected by the competitor.}
\label{fig:final result}
\end{figure*}

\textbf{SLV} \textit{vs.} \textbf{Other labeling schemes.}
Regarding SVL as a pseudo labeling strategy, we compare 3 different labeling schemes and analyze the strengths and weaknesses of them respectively.
The first scheme is a conventional version that selects the highest-scoring proposal for each positive class.
The second scheme is a clustering version that selects the highest-scoring proposal from every proposal cluster for each positive class.
And the last scheme is the proposed SLV. Fig \ref{fig:label} contains a few labeling examples of 3 schemes in different scenarios,
the first row shows that the SLV module is beneficial to find as many labels as possible rather than only one for each positive class.
Then, the second row shows the property of 3 schemes when labeling larger objects and the bounding boxes labeled by SLV have higher $IoU$ with ground-truth boxes.
However, as shown in the third row of Fig \ref{fig:label}, when objects are gathering closely, the SLV is prone to labeling these
objects as one instance. Meanwhile, all 3 schemes failed when labeling the “table” due to its weak feature representation
(the plate in the table is labeled instead). This is an issue worth exploring in future work. Despite these bad cases, the performance
of the network with SLV (53.5\% mAP) still surpasses its counterparts using two other labeling schemes (52.1\% mAP for the first scheme and 52.4\% mAP for the second scheme).

\subsection{Comparison with Other Methods}
In this subsection, we compare the results of our method with other works.
We report our experiment results on PASCAL VOC 2007 and 2012 datasets on Table \ref{mAP-2007}, Table \ref{corloc-2007} and Table \ref{results-2012}.
Our method obtains 53.5\% on mAP and 71.0\% on CorLoc with single VGG16 model
on VOC 2007 dataset, which outperforms all the other single model methods. We further re-train a
Fast-RCNN detector based on pseudo ground-truths produced by SLV (VGG) and the re-trained
model obtains 53.9\% on mAP and 72.0\% on CorLoc on VOC 2007 dataset, which are the new state-of-the-arts.
On VOC 2012 dataset, our method obtains 49.2\% on mAP, which is also the best in all the single model methods  and obtains 69.2\% on CorLoc.

Different from the recent works, \eg\cite{yang2019towards}, that select high-scoring proposals as pseudo ground-truths to enhance
localization ability, the proposed SLV is devoted to searching the boundaries of different objects from a more macro perspective
and thus obtains a better detection ability.
We illustrate some typical detection results of our method and a competitor model in Fig \ref{fig:final result}.
It is obvious that the bounding boxes output by our method have a better localization performance.
This is due to our multi-task network is able to classify and localize proposals at the same time, while the competitor is single-task form and only highlights the most discriminative object parts. Though our method outperforms the competitor
significantly, it is also worth noting that the detection results on some classes like “chair”, “table”, “plant” and “person”, are undesirable sometimes 
(last row of Fig \ref{fig:final result}).
We suggest that the supervisions generated in SLV module are not precise enough in object-gathering
scenarios: many chairs are gathering or an indoor table surrounded by many other objects.


\section{Conclusion}
In this paper, we propose a novel and effective module, spatial likelihood voting (SLV), for weakly supervised object detection.
We propose to evolve the instance classification problem in most MIL-based models into multi-tasking field
to shorten the gap between weakly supervised and fully supervised object detection.
The proposed SLV module converges the proposal localizing process without any bounding box annotations and
an end-to-end training framework is proposed for our model. The proposed framework obtains better classification and localization performance through end-to-end multi-tasking learning. 
Extensive experiments conducted on VOC 2007 and 2012 datasets show substantial improvements of our method compared with previous WSOD methods.


\section{Acknowledgements}
This work was supported by the Fundamental Research Funds for the Central Universities,
and the National Natural Science Foundation of China under Grant 31627802.

{\small
\bibliographystyle{ieee_fullname}
\bibliography{egbib}
}

\end{document}